\def\year{2022}\relax
\documentclass[letterpaper]{article} % DO NOT CHANGE THIS
\usepackage{aaai22}  % DO NOT CHANGE THIS
\usepackage{times}  % DO NOT CHANGE THIS
\usepackage{helvet}  % DO NOT CHANGE THIS
\usepackage{courier}  % DO NOT CHANGE THIS
\usepackage[hyphens]{url}  % DO NOT CHANGE THIS
\usepackage{graphicx} % DO NOT CHANGE THIS
\usepackage{natbib}  % DO NOT CHANGE THIS AND DO NOT ADD ANY OPTIONS TO IT
\usepackage{caption} % DO NOT CHANGE THIS AND DO NOT ADD ANY OPTIONS TO IT

\DeclareCaptionStyle{ruled}{labelfont=normalfont,labelsep=colon,strut=off} % DO NOT CHANGE THIS

\usepackage{amsmath}
\usepackage{algorithm}
\usepackage{algorithmic}
\usepackage[scr=rsfs]{mathalpha}

\usepackage[textsize=scriptsize,textwidth=2.3cm,shadow]{todonotes}
\usepackage{fdsymbol}

\usepackage{newfloat}
\usepackage{listings}
\floatstyle{ruled}
\newfloat{listing}{tb}{lst}{}
\floatname{listing}{Listing}
\title{Finding Counterfactual Explanations through Constraint Relaxations}
\author{
    Sharmi Dev Gupta\textsuperscript{\rm1}\textsuperscript{\rm 4},
    Beg\"{u}m Gen\c{c}\textsuperscript{\rm2}\textsuperscript{\rm 4}, and
    Barry O'Sullivan\textsuperscript{\rm 1}\textsuperscript{\rm 2}\textsuperscript{\rm 3}\textsuperscript{\rm 4}
}
\affiliations{
   SFI Centre for Research Training in AI\textsuperscript{\rm 1}, 
    Insight Centre for Data Analytics\textsuperscript{\rm 2}, 
    Confirm Centre for Smart Manufacturing\textsuperscript{\rm 3}, 
    School of Computer Science \& IT, University College Cork, Ireland\textsuperscript{\rm 4}
    
    sharmi.devgupta@cs.ucc.ie,
    begum.genc@cs.ucc.ie,
    b.osullivan@cs.ucc.ie
}

\usepackage{bibentry}
\usepackage{xcolor}
\usepackage{ulem}

\usepackage{amsthm}
\newtheorem{definition}{Definition}
\begin{document}

\maketitle

\begin{abstract}
% Many times during dealing with Constraint Satisfaction Problems, we encounter infeasible solutions.
Interactive constraint systems often suffer from infeasibility (no solution) due to conflicting user constraints. 
% Either the constraints conflict with each other or the domain sizes conflict with constraints. 
%In any of the cases the only way to arrive at a solution is to eliminate certain constraints completely so as to get rid of conflicts and attain consistency.
A common approach to recover infeasibility is to eliminate the constraints that cause the conflicts in the system.
% and attain consistency.
%This approach lacks opportunity for the system to explain to users what changes in their constraint set can help in attaining the desired outcome. 
This approach allows the system to provide an explanation as: ``if the user is willing to drop out some of their constraints, there exists a solution''.
However, one can criticise this form of explanation as not being very informative.
A counterfactual explanation is %an approach
a type of explanation that can provide a basis for the user to 
%analyse their preferences in order to relax their constraints instead of the standard elimination process. 
recover feasibility by helping them understand which changes can be applied to their existing constraints rather than removing them.
This approach has been %researched a lot
extensively studied
in the machine learning field, but requires a more thorough investigation %when it comes to CSPs. 
in the context of constraint satisfaction. 
%Counterfactual explanations give the users not only the explanation of why the solution a set of given constraints is not feasible, but they also provide them options which might provide them a feasible solution. 
%This paper defines a counterfactual explanation in terms of problem and builds on some earlier work done in the machine learning domain in order to provide an algorithm suited particularly for constraint satisfaction problems.
We propose an iterative method based on conflict detection and maximal relaxations in  over-constrained constraint satisfaction problems to help compute a counterfactual explanation.

\end{abstract}

\section{Introduction}
%\bccom{Sharmi, I applied many changes on Abstract and Introduction. Please proof-read and correct where necessary.}
In the long-standing history of constraints, an \textit{explanation} often strives to interpret the reasons for an infeasible scenario. 
This interpretation mostly depends on the identification of minimal conflicts 
 (or minimal unsatisfiable subsets).
Conflicts have been studied extensively in areas such as model-based diagnosis, 
 Boolean satisfiability, product configuration, solving logic puzzles, interactive search,
 etc., where the user constraints play an important role~\cite{DBLP:conf/ijcai/GuptaGO21}.
For instance, in the context of model-based diagnosis, if an observed outcome is not 
 what was expected, then the goal is to provide an explanation to help understand \textit{which} sets of conditions led to that unexpected outcome.
Similarly, when solving a %constraint satisfaction problem such as the 
 scheduling problem, an explanation can provide insights to why the given problem is not feasible under the provided sets of background and foreground constraints, and removing which set of constraints can provide a relaxation to the problem such that one can find a feasible solution.
However, it is important to note here that these explanations are not always produced for the user, but sometimes produced for speeding up the search or debugging for the developer.

%A counterfactual explanation definition as used for interpreting machine learning models aims at providing explanations to the outcomes by changing the values of a number of input variables to generate an alternative outcome~\cite{benk2020explaining}. 
Recently, the need for user-centered explanations in AI has substantially increased due to 
 several important factors such as the black-box nature of complex AI applications, the
 \textit{right to explanation of a decision} in the EU's General Data Protection Regulations~\cite{gdpr}, 
 and the development of Trustworthy AI for building trust between AI and the society~\cite{ec2019ethics}.
To address this issue, Wachter et al. 
%highlight the difference between the typical form of explanations vs. counterfactual explanations as follows: ``Counterfactual explanations describe a dependency on the external facts that led to that decision, whereas the common form of explanation attempts to convey the internal state of an algorithm that led to a decision''~\cite{wachter2017counterfactual}.
 proposed to use counterfactuals from philosophy, and adapt them to the AI domain to explain
 algorithmic decisions~\cite{DBLP:journals/corr/abs-1711-00399,wachter2018counterfactual}.
They describe a \textit{counterfactual explanation} as a statement that explains the minimal
 change to the system that results in a different outcome.
By providing counterfactual explanations, %one can
 it is expected to improve the understandability of the underlying model, and support
 decision-making process of the user.
%They also describe an ideal counterfactual explanation as one that alters the values of the variables as little as possible to change the outcome.

A counterfactual explanation seeks to provide a minimal explanation to a 
 question of the form: ``Why is the outcome $X$ and not $Y$?''~\cite{wachter2018counterfactual}.
%In context of Constraint programming the approach towards XAI and more specifically counterfactual explanation varies slightly from the Machine learning approaches. In Constraint Programming the counterfactual explanations are not ambiguous and do not go through a black box instead the constraints and the conflicts are very transparent so it is easier to understand exactly what elements are required to change in order to arrive at a desired result. 
%This counterfactual approach in CP can very well be used for configuration and scheduling problems for example consider a manufacturer placing several orders for item deliveries, specifying priority on each order. After seeing the initial schedule, the manufacturer wants an explanation. They ask “Why was the schedule not different? Why is order A not delivered in two days, and B in five days?” \cite{korikov2021counterfactual-2}
To illustrate, consider a constraint system that aims to solve the course timetabling problem at
 a university. 
The dedicated admin staff runs the timetabling system to obtain a feasible timetable.
However, a lecturer, who is used to teaching their assigned course on Mondays, asks the admin:
 ``Why is my Course A scheduled to Friday instead of Monday? I cannot attend lectures on Fridays due to travel.''.
In order to accommodate this user constraint, which was not a part of the system before, 
 the admin can add this new information to the system as a background constraint to ensure it 
 is not violated.
However, adding the new constraint may cause an infeasible state in the system.
To recover from this situation, the admin can follow a traditional conflict elimination mechanism,
 which involves finding a set of constraints to relax so the conflicts in the problem
  are removed.
Alternatively, the system can provide a counterfactual explanation that explains: ``If you move
 Course B from Monday to Tuesday, you can schedule Course A on Friday.''.
Note that, if the user's request does not cause an infeasibility, alternative explanations can
 be considered such as: ``Given the new constraint, an alternative schedule can be found at an extra cost of X.''.
%But just an explanation of why the desired outcome was not achieved is not very useful to the user as it does not help him understand how he could achieve the desired outcome. We try to bridge this gap through our novel approach.Our goal in this paper is to define a counterfactual explanation and to build on this concept of maximal relaxation of the counterfactual explanation through our novel algorithm and render consistency towards an infeasible CSP.
 
Counterfactual explanations have recently been adapted to optimization problems~\cite{DBLP:conf/ijcai/KorikovSB21}.
%In Section~\ref{sec:relevantwork} 
We discuss relevant work in more detail in the Related Work section.
We then %in Section~\ref{sec:methodology} 
 propose a new approach to finding a counterfactual explanation based on identifying conflicts and maximal relaxations, demonstrate our model on a configuration problem, and conclude with a discussion and identification of some future directions.  

\section{Related Work}
\label{sec:relevantwork}
%Constraint Programming has many applications in scheduling, resource allocation, product configuration, etc. 
Our work focuses on explanations in the constraint satisfaction branch of AI working with a multi-point relaxation system. 
Infeasibility in constraint systems may cause an enormous cost at an industrial level, which includes customer dissatisfaction. 
Hence, explanation generation has been a very active and interesting topic.
The existing work on this topic has mostly focused on identification conflicts
 in the constraint satisfaction literature and also other relevant areas such as Boolean satisfiability~\cite{DBLP:conf/ijcai/GuptaGO21,DBLP:conf/ijcai/0001M20}. 
%\sharmi{discuss concept and papers(related works) maximal relaxations,conflicts,quickxplain, quantifiedxplain}
%CSPs are important for applications like 
%\bccom{This paper is not a survey paper, so we are not expected to explain every paper in detail, but we must show here how the field is advancing, who is proposing something for the first time, who is following it, what are the connections, what are all relevant papers, etc. I am making quite a lot of modifications to this section, but I think there is more work to do here.}
In this paper, we propose to adapt \textit{counterfactual explanations} to constraint-based systems. 
Up to date, counterfactual explanations are mostly studied under the XAI branch of
 machine learning systems and attracted a lot of attention.
%\subsection{Counterfactual Explanations}

In 2017, Wachter et al. proposed to use counterfactual explanations as a way to
 provide a minimal amount of information capable of altering a decision without understanding the internal logic of a model~\cite{DBLP:journals/corr/abs-1711-00399,wachter2018counterfactual}. 
In a recent survey paper on counterfactuals in XAI, Keane et
 al.~\shortcite{DBLP:conf/ijcai/KeaneKDS21} presented a detailed analysis of 100 distinct counterfactual methods and their overall evaluation and shortcomings along with a roadmap to improvement.
%Maximilian Schleich et al. present GeCo the first interactive system to compute feasible counterfactual explanations in real time.It relies on genetic algorithm which is customised to favour searching counterfactual explanations which require minimum changes. The authors also introduce two optimization methods which speeds up the genetic algorithm namely delta-representation and partial evaluation. The authors compared the results obtained from GeCo system with the results from 5 other state of the art systems and concluded that GeCo gave better quality results in real time\cite{DBLP:journals/pvldb/SchleichGZS21} .
They highlighted that only a few of these approaches are supported by user evaluations. 
Similarly, Miller argued that in explainable AI, a `good explanation' is usually
 defined by the researchers, but the social science dimension to this definition is not explored well~\cite{MILLER20191}.
Miller characterised explanations as \textit{contrastive}, \textit{selected}
 in a biased manner, \textit{social} (i.e. transferring knowledge), and not completely based on \textit{probabilities} (the most likely explanation is not necessarily the best explanation).  

Explanation generation in constraint satisfaction is usually achieved by identification
 of minimal conflicts (or minimal unsatisfiable subsets), or maximal relaxations~\cite{junker2001quickxplain,DBLP:journals/jar/LiffitonS08,DBLP:conf/aaai/OSullivanPFP07}.
%present an algorithm to compute the generalisation of conflict based explanations of inconsistency for a Quantified Constraint Satisfaction Problem.
%The approach followed in this paper is similar to our approach since it uses the Quickxplain algorithm and modifies it such that the removal of a constraint is replaced by a specific requirement relaxation\cite{ferguson2007quantified}.
Despite the long-standing history of explanation generation in
 constraint satisfaction, the notion of counterfactual explanations is a relatively
 new and interesting topic.
However, there exist a few relevant studies that discuss related notions such as
 contrastive and abductive explanations in Boolean satisfiability.
 %Counterfactuals come into play here as an opportunity for resolving multiple conflicts in a cost-efficient manner. 
%\bccom{I don't think we can claim cost-efficiency yet as we haven't studied it. So, I have removed the above sentence.}
%Various works have been done in the past in Conflict detection for CSPs and Counterfactual explanations (especially for machine learning algorithm).
As an example, Ignatiev et al. have a number of studies at the intersection of ML and SAT~\cite{DBLP:conf/cade/IgnatievPNM18, ignatiev2020relating}.
Their work %builds a connection between the machine learning and unsatisfiable formulas by 
discusses different types of explanations, such as \textit{local abductive} (answering ``Why prediction X?'') and \textit{contrastive} explanations (answering ``Why not?'').
% and focuses on the relation between local abductive and contrastive explanations~\cite{ignatiev2020relating}. 
More specifically, the authors discuss how recent approaches for computing abductive
 explanations can be exploited for computing contrastive/counterfactual explanations. 
Their findings highlight an important property that the model based local abductive and
 contrastive explanations are related by minimal hitting set relationships~\cite{ignatiev2020relating}.
%\bccom{Counterfactual-relevant work starts with the Ignatiev's paper here. The previous ones are on conflict detection.}
%Their study shows how recent approaches for computing abductive explanations can be exploited for computing contrastive/counterfactual explanations.% The paper also demonstrates that model based local abductive and contrastive explanation are related by minimal hitting set relationships.
%Their study demonstrates that a local abductive and contrastive explanations are related by a minimal hitting set relationship.
More recently, Cooper and Marques-Silva investigate the computational complexity of finding a subset-minimal abductive or contrastive explanation of a decision taken by a classifier~\cite{cooper2021tractability}.  
The authors define the explanation notions analogous to Ignatiev et al.~\cite{ignatiev2020relating}.

%\bccom{Ignatiev paper has been discussed above. Can you merge?}
In parallel, Cyras et al. present an extensive overview of various machine reasoning techniques employed in the domain of XAI, in which they discuss XAI techniques from symbolic AI perspective~\cite{cyras2020machine}.
%They mention how contrastive explanations differ from counterfactual explanations and how contrastive explanation can work via counterfactuals. 
The authors classify explanations into three categories.
These are namely \textit{attributive}, \textit{contrastive}, and \textit{actionable} explanations. 
Subsequently, they discuss the links between these explanation notions and the existing notions in symbolic AI by covering many different topics such as abductive logic programming, answer set programming, constraint programming, SAT, etc.
They discuss that contrastive explanations can be achieved via counterfactuals and define a \textit{counterfactual contrastive explanation} as ``making or imagining different choices and analysing what could happen or could have happened''.
On the other hand, they define an actionable explanation as one that aim to answer ``What can be done in order for a system to yield outcome $o$, given information $i$?''.
%The authors mention that contrastive explanation can be seen as an extension of attributive explanations and that counterfactual explanations continue to work in an end to end fashion. A contrastive counterfactual explanation would entail imagining a more desirable outcome and addresses the changes required in the system to get to that outcome.

%\sharmi{add more papers}
%\sharmi{check this one again}Martin C. Cooper and João Marques-Silva investigate the computational complexity of providing a correct and minimal explanation for a decision made by different families of classifiers. They take their motivation from the question of which classifiers can allow formal and correct explanations to be computed in polynomial time\cite{cooper2021tractability}. 
%\sharmi{I cant understand the above paper}

%\bccom{There is the recent CP 2021  paper from Beck on counterfactuals: Counterfactual Explanations via Inverse Constraint
%Programming. They also have a previous one on GDPR. These two papers must be discussed and we must explain how out work is different from theirs.}
%\sharmi{please check this para}\\
%Korikov et al. develop an approach to generate counterfactual explanation for optimization based decisions using general inverse combinatorial optimization.
%\sharmi{added. Please check}\\
Explanation generation is also quite important for predictive models for they must provide justification for their decision along with alternative solutions specifically solutions which are closest to the user requirements. There have been some recent novel work on generating the nearest counterfactual explanation; Amir-Hossein Karimi et al. present a model agnostic, data type agnostic and distance agnostic algorithm which is able to generate plausible and diverse counterfactual explanations for any sample data. Their model generates counterfactuals at more favourable distances compared to recent optimization based approaches and also informs system administrators about the potential unfair dependence of the model on certain protected attributes\cite{pmlr-v108-karimi20a}. %\bccom{Sharmi, it is not clear to me how this paragraph connects with the rest of the text. Can you connect it with the logically relevant part and build the connection? }

To the best of our knowledge, the most relevant study to our work has recently been conducted by Korikov et al., in which the authors extend the notion of counterfactual explanations to optimisation-based decisions by using inverse optimisation~\cite{DBLP:conf/ijcai/KorikovSB21}.
They assume that the user is interested in an explanation of why a solution to an optimisation problem does not satisfy a set of additional user constraints that were not initially expressed by the user.
In their work, the authors define counterfactual explanations analogous to those of Wachter et al.~\cite{wachter2018counterfactual}.
They aim to find the \textit{nearest counterfactual explanation}, which corresponds to finding a set of changes on the features such that the new solution is as close to the previous one as possible.
%\bccom{Have you read the related work section for this paper? They mention some counterfactual work in planning. If not, can you please check? These are all works that you should be familiar with.}
The authors also highlight that the links between conflict-detection mechanisms in constraint satisfaction and counterfactual explanations is not clear. 
%\bccom{Which work of Korikov? + Korikov has co authors so should be cited as Korikov et al. Are you referring to Karimi's work you described above?}
%They generate counterfactual explanation for general optimization model with two common objective function. The methods presented in the paper can be used to produce user centred explanations and gives user the opportunity to contest and change decision of the system. This also makes it relevant to the GDPR guidelines that mandate right to explanation in XAI\cite{korikov2021counterfactual-1}. 
%Subsequently, Korikov and Beck present in their paper how a counterfactual explanation can be found using inverse optimisation in constraint programming.
Subsequently, Korikov and Beck generalize their work to constraint programming and show
 that counterfactual explanations can be found using inverse constraint programming using a cost vector~\cite{korikov2021counterfactual-2}. 
 Karimi \cite{pmlr-v108-karimi20a} along with Korikov ~\cite{DBLP:conf/ijcai/KorikovSB21}  have a similar goal to generate the optimal counterfactual explanations for classifiers. Karimi  however does not take into account decisions taken by explicit optimization models as opposed to Korikov.
%Their method aspires to find the nearest counterfactual explanation for optimisation problems and user queries using inverse constraint programming. The drawback of their method is however they cannot find the nearest counterfactual explanation for set of user constraints described by more expressive constraints. 
%The author's approach takes any optimal solution available to it without being able to express the relevance and the reasoning behind the changes to the user. 
%On the contrary our approach uses multi point relaxation to achieve feasibility and can explain to the user the exact variables required for the changes and the exact difference between their original and relaxed values \cite{korikov2021counterfactual-2}. 

In this paper, our goal is to find a counterfactual explanation to a given constraint
 problem by using conflicts and constraint relaxation, and address the question that Korikov et al. raised related to the connection between conflicts and counterfactuals~\cite{DBLP:conf/ijcai/KorikovSB21}.
To achieve this, we use a relevant work from Ferguson and O’Sullivan as the foundation
 of our proposed method, in which the authors generalize conflict-based explanations to Quantified CSP framework~\cite{ferguson2007quantified}.
Their approach extends the famous \textsc{QuickXplain} algorithm~\cite{Junker2004} by
 allowing relaxation of constraints instead of their removal from the constraint set.
We also demonstrate how this mechanism based on identification of maximal relaxations
 can be used to find counterfactual explanations in constraint-based systems.

\section{Methodology}
\label{sec:methodology}
First, we define some important notions existing in the Constraint Programming literature on explanations, define counterfactual explanations, and discuss the relation with a counterfactual explanation and constraint relaxation. 
Consequently, we present our proposed model to find a counterfactual explanation and demonstrate it on a sample item configuration problem.
 
\subsection{Preliminaries}

A constraint satisfaction problem (CSP) is defined as a 3-tuple 
 $\mathcal{\phi} := (\mathcal{X}, \mathcal{D}, \mathcal{C})$ where 
 $\mathcal{X} := \{ x_1, x_2, ..., x_n \}$ is a finite set of variables, 
 $\mathcal{D} := \{ D(x_1)$, $D(x_2), ..., D(x_n) \}$ denotes the set of finite domains 
 where the domain $D(x_i)$ is the finite set of values that variable $x_i$ 
 can take, and a set of constraints 
 $\mathcal{C} := \{c_1, c_2, ..., c_m\}$. 
More specifically, a problem $\mathcal{\phi}$ in Constraint Programming 
 can be defined using two sets of constraints $\mathcal{B}$ representing the \textit{background constraints} and $\mathcal{F}$ representing the 
 \textit{foreground constraints} (or \textit{user requirements/constraints}) 
 in the context of configuration problems or other interactive settings. 
Using this alternative representation, a problem is notated as 
 $\mathcal{\phi} := (\mathcal{X}, \mathcal{D}, \mathcal{C})$, where 
 $ \mathcal{C} := \mathcal{B} \cup \mathcal{F}$.
In order to increase readability, we sometimes refer to a problem as 
 $P :=(\mathcal{B}, \mathcal{F})$.
A set of constraints is called \textit{inconsistent} (or \textit{unsatisfiable})
 if there is no solution. 
In this case, the problem is said to be \textit{infeasible}.
If the problem has at least one solution, the set of constraints is said to be
 \textit{consistent} (or \textit{satisfiable}), and the related problem is 
 referred to as \textit{feasible}.
We assume that the set of background constraints are consistent, but the user
 constraints may introduce infeasibility.
%A set of constraints $C$ is referred as over-constrained if they cannot be satisfied at the same time.
%\bccom{There are two definition for a CSP. The above and below paragraph. Which one should we take into account?}
We define below a number of relevant definitions existing in the literature.

%\begin{definition}
%[Classical Constraint Satisfaction Problem]
%A constraint satisfaction problem (CSP) is defined as a 3-tuple $\mathcal{P} \hateq (\mathcal{X}, \mathcal{D}, \mathcal{C})$ where $\mathcal{X}$ is a finite set of variables $\mathcal{X} \hateq (x_1, x_2, ..., x_n) \mathcal{D}$ is a set of finite domains $\mathcal{D} \hateq (D(x_1)$, $D(x_2), ..., D(x_n))$ where the domain $D(x_i)$ is
%the finite set of values that variable $x_i$ can take, and a set of
%constraints $\mathcal{C} \hateq (c_1, c_2, ..., c_m)$~\cite{ferguson2007quantified}. 
%Each constraint $c_i$ is defined
%by the ordered set var($c_i$) of the variables it involves, and a
%set sol($c_i$) of allowed combinations of values. An assignment
%of values to the variables in var($c_i$) satisfies $c_i$ if it belongs
%to sol($c_i$). A solution to a CSP is an assignment of a value
%from its domain to each variable such that every constraint
%in C is satisfied.
%\bccom{There is a lot of unnecessary notation in this definition. We do not need most of them.}
%\end{definition}

\begin{definition}[Conflict~\cite{Junker2004}]
A subset $C$ of $\mathcal{F}$ is a conflict of a problem $P :=
(\mathcal{B}, \mathcal{F})$ iff $\mathcal{B} \cup C$ has no solution.
\end{definition}

\begin{definition}[Minimal Conflict~\cite{Junker2004}] 
A conflict $C$ of $\mathcal{F}$ is minimal (irreducible) if each proper subset of $C$ is consistent with the background $\mathcal{B}$ (or if no proper subset of $C$ is a conflict).
\label{def:minimalconflictset}
\end{definition}

\begin{definition}[Relaxation~\cite{Junker2004}]
A subset $R$ of $\mathcal{F}$ is a relaxation of $P := (\mathcal{B}, \mathcal{F})$ iff $\mathcal{B} \cup R$ has a solution.
\end{definition}

\begin{definition}[Maximal Relaxation~\cite{DBLP:conf/aaai/OSullivanPFP07}]
A subset $R$ of $\mathcal{F}$ is a maximal relaxation of a problem
 and there is no $\{c\} \in \mathcal{F} \setminus R$ such that 
 $\mathcal{B} \cup R \cup \{c\}$ also admits a solution.
\end{definition}

A problem is said to be \textit{over-constrained} if it contains an
 exponential number of conflicts and an exponential number of relaxations.
Based on the definition of a maximal relaxation, the complementary notion of 
 minimal exclusion set can be defined.

\begin{definition}[Minimal Exclusion Set%~\cite{6984549}
~\cite{DBLP:conf/aaai/OSullivanPFP07}] 
Given a problem $P := (\mathcal{B}, \mathcal{F})$ that is inconsistent, 
 and a maximal relaxation $R \subseteq \mathcal{F}$, 
 $E = \mathcal{F} \setminus R$ denotes a minimal exclusion set.
%A set of constraints $C' \subseteq C$ is a minimal exclusion set of constraints of $C$ iff $P($C$ \setminus $C'$)$ is feasible and for every $c \in C'$
%, $P($C$ \setminus (C' - {c}))$ is unfeasible.
\end{definition}

Note that, the definitions above are defined under \textit{two-point relaxation spaces}.
A two-point relaxation space either allows to have the constraint in the constraint set, or not.
In this paper, we work under \textit{multi-point relaxation spaces}, 
 which correspond to replacing a constraint with any weaker one~\cite{ferguson2007quantified,DBLP:conf/ictac/MehtaOQ15}. 
To illustrate this, consider the user constraint in Equation~\ref{eqn_constraint} between two variables. 

\begin{equation}
 x_1 \in \{1, 2, 3\}, x_2 \in \{3, 4\}.\{x_1 > x_2\}
\label{eqn_constraint}
\end{equation}

Equation~\ref{eqn_constraint} is an inconsistent constraint. 
Assuming that all remaining constraints are consistent, one
 can remove this constraint from the constraint set to recover consistency in a
 two-point relaxation space.
Alternatively, in a multi-point relaxation space, this constraint can be
 relaxed to Equation~\ref{eqn_constraint_relaxed}, which evaluates to
 \textsc{True} as there exist satisfying values: $x_1 = 3, x_2 = 3$.
We say that Equation~\ref{eqn_constraint} is a \textit{tighter} version of
 Equation~\ref{eqn_constraint_relaxed}, and the Equation~\ref{eqn_constraint_relaxed}
 is a \textit{relaxed} version of the former.

\begin{equation}
x_1 \in \{1, 2, 3\}, x_2 \in \{3, 4\}.\{x_1 \geq x_2\}
\label{eqn_constraint_relaxed}
\end{equation}

\subsection{Finding a counterfactual explanation in CSP}

We define a counterfactual explanation by adapting the definitions from 
 Wachter et al.~\shortcite{wachter2018counterfactual} and Korikov et al.~\shortcite{DBLP:conf/ijcai/KorikovSB21}.
We aim to find an explanation to the user with minimal changes to her constraints that
 informs the user on how to recover from an infeasible state.
In other words, given a problem $P := (\mathcal{B}, \mathcal{F})$,
 and a user constraint $\{c\} \not\in \mathcal{F}$ and 
 $P' := (\mathcal{B}, \mathcal{F} \cup \{c\})$ is infeasible, we define 
 a \textit{counterfactual explanation} as a set of constraints that 
 explain the minimal set of changes in
 $\mathcal{F}$ so that the problem $P'$ with the updated constraints
 becomes feasible.
In Definition~\ref{def:CE} we formally define to a counterfactual explanation
 based on maximal relaxations in CSP.

\begin{definition}
Define two CSPs as 
 %$\mathcal{\phi} = \langle \mathcal{X}, \mathcal{D}, \mathcal{C} \rangle$ with $\mathcal{C} := (\mathcal{B}, \mathcal{F} \cup \{c\})$
 $P := (\mathcal{B}, \mathcal{F} \cup \{c\})$ that is inconsistent and %$\mathcal{\phi}' = \langle \mathcal{X}, \mathcal{D}, \mathcal{C'} \rangle$ with $\mathcal{C'} := (\mathcal{B} \cup \{c\}, \mathcal{F}')$ that is consistent,
 $P' := (\mathcal{B} \cup \{c\}, \mathcal{F'} )$ that is consistent,
 where a constraint $\{c\} \not\in \mathcal{C}$ denotes a counterfactual
 user constraint, and $\mathcal{F}'$ corresponds to a minimal set of changes
 applied to $\mathcal{F}$ such that $P'$ becomes consistent.
A counterfactual explanation, denoted by $\mathcal{E}$, corresponds to a
 minimal set of changes required on user constraints to change the state of
 the problem, where $\mathcal{E} = \mathcal{F}' \setminus \mathcal{F}$.
 \label{def:CE}
\end{definition}

Observe that, this system can be generalized to any infeasible problem
 $P := (\mathcal{B}, \mathcal{F})$ to explain how to recover
 feasibility without requiring any counterfactual user constraint.

\begin{figure}[t]
\centering 
\includegraphics[width=0.25\textwidth]{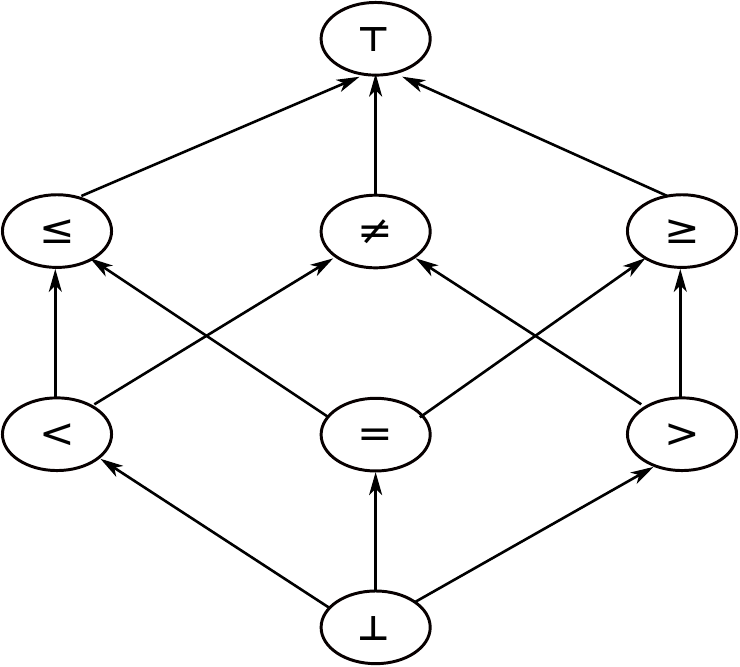}
\caption{A sample partially ordered set of states for numerical constraints in a multi-point relaxation space.}
\label{fig:int_dom_lattice}
\end{figure}

Our method assumes the existence of a multi-point relaxation space defined
 by the knowledge engineer for each variable in the problem. 
The relaxation space of a feature may take different characterisations, such as 
 a partially ordered set, lattice, hierarchical ordering, etc. 
Using these structures pave the way to have comparable or incomparable relaxation states. 
A top element $\top$ must be defined for each relaxation space, which corresponds
 to maximally relaxing the relevant constraint (eliminating from the constraint set).
Similarly, a bottom element $\perp$ denotes an infeasible state for a given
 constraint.
To illustrate, Figure~\ref{fig:int_dom_lattice} can be considered as a
 multi-point relaxation space for equality or inequality constraints that deal
 with numerical variables.
For instance, given an equality constraint such as $x = 5$, the constraint can
 be relaxed to $x \leq 5$ or $x \geq 5$, where the two states are incomparable 
 on the partially ordered set of states.
For the sake of notation, we denote comparable states as 
 $\{ \top \} \sqsubseteq \{ \leq \} \sqsubseteq \{=\} \sqsubseteq \{\perp\}$, where
 $\{ \top \} \sqsubseteq \{ \leq \}$ is read as state $\{ \top \}$ \textit{dominates} state $\{ \leq \}$.
%If a constraint is further relaxed from that state, it corresponds to having no constraints, i.e. $\top$.

%\bccom{We have decided to take the relaxation space from the data in the last meeting. Instead of this image, can you formally explain the procedure here?}
%\sharmi{I have a lot of doubt regarding the relaxation space generated from the data set now. Just go through it once and we can discuss if it works or not}
%We use a multi point relaxation method in our algorithm in order to iterate through all the different relaxation spaces for each constraint. A multi point relaxation space corresponds to various alternatives for the constraints relaxed bit by bit.Every constraint would have its own relaxation space.

%\sharmi{This is what I am thinking about fixing in the algorithm}\\
%\sharmi{retrieve some products from the dataset where }\\
%\sharmi{Each relaxation space $\phi'$ would consist of infeasible values taken from the data set since they would all be relaxation of the constraint. So  $[ \mathcal{R}_1, \ldots, \mathcal{R}_n]$ would be taken from the dataset. Let us say my price constraint is $<$500, my relaxation space would consist of all values greater than or equal to 500}\\
 %\sharmi{Start with the least relaxed values in all the relaxation space and compare it with the dataset to find a matching product.If no product found take one relaxation space at a time and relax it a bit more and recombine and check for a match in the data set. If a match is found we compare the values in each relaxation space to the original constraint. If it is not equal then it is a counterfactual explanation and the new value is maximally relaxes }\\

\begin{algorithm}[t]
\caption{CounterFactualXplain ($\phi, \mathcal{R}$)}
\begin{algorithmic}
\STATE \textbf{Input:} A CSP $\phi = \langle \mathcal{X}, \mathcal{D}, \mathcal{C}\rangle$, where $ \mathcal{C} =  \mathcal{B} \cup  \mathcal{F}$, a set of multi-point relaxation spaces of each user constraint $\mathcal{F} = [c_1, \ldots, c_n]$ as $\mathscr{R} = [\mathcal{R}_1, \ldots, \mathcal{R}_n]$.

\STATE \textbf{Output:} A counterfactual explanation $\mathcal{E}$, and a maximal relaxation $C'$.
\STATE $n = \vert \mathcal{F} \vert$, $\mathcal{E} = \emptyset$, $C' = \emptyset$ 
\IF { $\phi$ is feasible} 
  \RETURN{ no conflict } 
\ENDIF
\IF {$\forall i \in \{1, \ldots, n \} \vert \mathcal{R}_i \vert = 1$} 
  \RETURN{ no relaxation }  
\ENDIF  

\STATE $C' := \mathcal{B} \cup \{ \top_i \vert \top_i$ is top in $\mathcal{R}_i, \forall i \in  \{1, \ldots, n \} \}$% \top_2, \ldots, \top_n\}$ for each $\{ \mathcal{R}_1, \ldots, \mathcal{R}_n \}$
\STATE $\phi' = \langle \mathcal{X}, \mathcal{D}, C' \rangle$

\FOR{$c_i \in \mathcal{F}$} 
    \STATE choose $r_j$ from maxima of $\mathcal{R}_i$ of $c_i$ s.t. $r_j \not\in C'$ and $r_j \subseteq c_i$ %corresponding to $c_i$
   %\STATE Start with $r_j$ =  $\top_i$ corresponding to each $c_i$
  %\STATE Compare $r_j$ with each item in relaxation space $[ \mathcal{R}_1, \ldots, \mathcal{R}_n]$ based on operators corresponding to $c_i$. \COMMENT{ If user constraint was 'at most' then $r_j$ will check for \leq $R_n$ and $<R_n$ }
    \WHILE{$C' \cup \{r_j\}$ is consistent}
    %while the $\phi'$ is feasible add restrict the relaxation space to just the data set that follows the product
        \STATE $C' = C' \cup \{r_j\}$ 
         
        %\STATE $r_{prev} := r_j$
      
        \IF{$r_j$ equals $c_i$ }
            \STATE break
        \ENDIF
        
        \STATE $r_{prev} := r_j$
        \STATE choose maximal $r_j$ from $\mathcal{R}_i$ such that $\{r_j\} \not\in C'$ and $\{r_{prev}\} \sqsubseteq \{r_j\}$ %corresponding to $c_i$
        %\ELSE
        %    \STATE remove $r_{prev}$ from $\mathcal{R}_i $        
        %    \STATE $r_j$ is the next maxima of $\mathcal{R}_i$
        %\ENDIF
    \ENDWHILE

    %\STATE $\mathcal{M} = \mathcal{M} \cup \{r_{prev}\}$ \COMMENT {$r_{prev}$ was the last constraint that was feasible} 
    \IF{$c_i \not= r_{prev} $}
        \STATE $\mathcal{E} = \mathcal{E} \cup \{r_{prev}\}$ \COMMENT{ $r_{prev}$ is a part of the explanation}
    \ENDIF
\ENDFOR
\RETURN $\langle \mathcal{E}, C' \rangle$
\end{algorithmic}
\label{alg:CFxplain}
\end{algorithm}

%For our algorithm we are trying to use a more linear approach to resolving the problem. %\bccom{It is preferred to write the text without itemizing in academic papers.}
%The algorithm takes a CSP as an input. An ideal output would be the counterfactual explanation along with maximal relaxation.
Algorithm~\ref{alg:CFxplain} presents our proposed method \textsc{CounterFactualXplain}.
This approach is an adaptation of the \textsc{QuantifiedXplain} algorithm
 that was proposed to solve Quantified CSPs following a set of different relaxation forms including single constraint relaxation, relaxation of existentially/universally quantified domain, quantifier relaxation, etc. ~\cite{ferguson2007quantified}. 
From the set of different relaxation forms they propose, we only adapt
 single constraint relaxations in our work. 
Our proposed method follows an iterative approach for identifying maximal
 relaxations of the problem. 
Note that, if the relaxation spaces are two-point (binary), then the algorithm
 becomes a version of Junker's \textsc{RePlayXplain} algorithm that is an iterative approach to find a minimal conflict~\cite{junker2001quickxplain}.% \bccom{Sharmi, can you confirm if this replayxplain is correct?}

The \textsc{CounterFactualXplain} admits a CSP $\phi$ and the multi-point 
 relaxation spaces of each constraint that can be relaxed, and returns a counterfactual
 explanation $\mathcal{E}$ (a set of constraints that needs to be changed to restore feasibility) alongside a relaxed and feasible version of the constraint set of $\phi$. 
If the given CSP is feasible, then the algorithm returns `no conflict'. 
%If there are no multi point relaxations available for the constraint then we return no relaxation. When both the above cases are not valid we go to the third and main section of the algorithm. 
Similarly, if there is no relaxation space defined for all foreground constraints, 
 the algorithm returns `no relaxation'. 
% We follow the QuantifiedXplain algorithm where each constraint is relaxed from the most to the maximal relaxation point such that the constraint is feasible with minimum change to it and the problem can arrive at a solution.
For any other problem, the algorithm creates a copy CSP $\phi'$ 
 with the original set of variables and domains, but uses a constraint set $C'$ that initially contains only the top elements of each relaxation space for each constraint in $\mathcal{F}$. 
%In other words, there are no constraints in $\phi'$, hence, there are no conflicts.
Then, the procedure iteratively attempts to \textit{tighten} the maximal relaxation of 
 each constraint until either the original user constraint is reached or 
 an inconsistent set of constraints is formed. 
In this context, tightening a constraint $c$ corresponds to adding a 
 more restrictive form of $c$ to the existing set of constraints. 
In the case of having incomparable states in the relaxation space, when 
 tightening a constraint, first a path from the top element to the original constraint
 is found.
Next, each path is explored from the most relaxed state to the tighter ones on the path.

\section{Demonstration}

Consider a small problem from the item configuration domain, in which a user
 wants to purchase a laptop.
Assume there exist five different properties for each laptop: brand, screen size,
 memory, battery life, and price. 
Table~\ref{table:dataset} lists all available laptops in the solution space.
Also assume that the knowledge engineer defines the relaxation spaces as directions
 for the numerical values (screen size, memory, battery life, and price) for this problem, and the brand relaxation space consists of incomparable states.  
There are two directions for the numerical values: MIB (``more is better'') and
 LIB (``less is better'').
Additionally, all brands are equally distant to each other.
The users are allowed to express their preferences on the direction of numerical 
 features.

\begin{table}[t]
\caption{The set of all available laptops.}
\label{table:dataset}
\centering
\footnotesize
\begin{tabular}{ |c|c|c|c|c|} 
\hline
Brand & Size (inches) & Memory (MB) & Life (hr) & Price\\ \hline
Lenovo & 15.4 & 1024.0 & 2.2 & $1499.99$\\ \hline
Sony & 11.1 & 1024.0 & 11.0 & $2349.99$\\ \hline
Lenovo & 15.0 & 512.0 & 10.0 & $2616.99$\\ \hline
HP & 15.0 & 512.0 & 4.5 & $785.99$\\ \hline
Lenovo & 14.0 & 512.0 & 4.5 & $1899$\\ \hline
\end{tabular}
\end{table}

For demonstration purposes, assume there exists a user who initially expresses 
 her preferred values for some of these properties.
In Table~\ref{fig-example2}, $c_1, c_2, c_3, c_4$ correspond to the initial constraints
 of the user.
The user is interested in finding a laptop with brand `Lenovo', screen size of at
 least 15 inches, memory of at least 512 MB, and battery life of at least 10 hours. 
The constraint system solves the problem, and returns the solution (item) to the
 user: \{Lenovo, 15.0 inches, 512.0 MB, 10 hr, \$2616.99\}. 
However, the user is not happy with the recommended item as she realises that the
 recommended item exceeds her budget. 
Therefore, she adds an extra constraint to the system by asking the 
 question: ``Why does the laptop recommended to me costs more than $\$2000$? I need
 an alternative that costs at most $\$2000$.''.
This user constraint is captured as $c_5$ in Table~\ref{fig-example2}.
Note that, we are interested in a solution that may not satisfy some user constraints but satisfies the counterfactual constraint. Therefore, we move the counterfactual constraint to the background constraints to avoid its relaxation by
 the \textsc{CounterFactualXplain} algorithm. %\sharmi{ There could be several scenarios which allow the user to interact with the system challenging certain solutions and specifying the cause. In this particular case the user is already unsatisfied with the cost and makes it clear that the cost must not exceed $\$2000$. In our consecutive iterations making this constraint a background constraint would ensure that the new solutions don't violate this background constraint }. 
 %\begum{I would go for a sentence: ``Note that, we are interested in a solution that may relax some user constraints but satisfies the counterfactual constraint. Therefore, we move this constraint.... (the prev. sentence)''} %\sharmi{ I don't understand this last line} \bccom{If we don't move the counterfactual constraint to background set, it may be relaxed. We don't want that.}

\begin{table}[t]
\caption{The list of user constraints ($c_1, c_2, c_3, c_4$) and the counterfactual constraint ($c_5$). The user preferences of directions are MIB (``more is better'') and LIB (``less is better'').}
\label{fig-example2} 
\centering
\footnotesize
\begin{tabular}{|cllc|}
\hline
$c_i$ & Property & User Constraint & Preference\\ \hline
$c_1$ & Brand & Lenovo & -- \\\hline
$c_2$ & Size (inches) & 15.0 & MIB \\\hline
$c_3$ & Memory (MB) & 512.0 & MIB\\\hline
$c_4$ & Life (hr) & 10.0 & MIB\\ \hline \hline
$c_5$ & Price & 2000 & LIB \\ \hline
\end{tabular}
\end{table}

% \begin{table}[t]
% \centering
% \footnotesize
% \begin{tabular}{cllc}
% \hline
% $c_i$ & Property & User Constraint & Preference\\ \hline
% $c_5$ & Price & 2000 & LIB \\
% \hline
% \end{tabular}
% \caption{A counterfactual constraint provided by the user in addition to the constraints provided in Table~\ref{fig-example2}.}
% \label{table:extra-constraint} 
% \end{table}

As our relaxation spaces are defined as directions, we use an ordered list representation.
Table~\ref{table:relaxation} presents the relaxation spaces for all constraints, where
 features are ordered with respect to the user's preference of direction.
If the user does not have a preference, we assume the direction is the default 
 direction provided by the knowledge engineer.

\begin{table}[t]
 \caption{Relaxation spaces for every feature for our data set.}
 \label{table:relaxation}
\centering
\footnotesize
\begin{tabular}{ |c|l| } 
\hline
  $c_i$ & Relaxation space of $c_i$ ($\mathcal{R}_i$) \\ \hline
  $c_1$ & $\top \subseteq$ \{HP, Lenovo, Sony\} $\subseteq \perp$\\\hline
  $c_2$ & $\top \subseteq 11.1 \subseteq 14.0 \subseteq 15.0 \subseteq 15.4 \subseteq \perp$ \\\hline
  $c_3$ & $\top \subseteq 512 \subseteq 1024 \subseteq \perp$ \\\hline
  $c_4$ & $\top \subseteq 2.2 \subseteq 4.5 \subseteq 10.0 \subseteq 11.0 \subseteq \perp$ \\\hline
  $c_5$ & $\top \subseteq 2616.99 \subseteq 2349.99 \subseteq 1899 \subseteq 1499.99 \subseteq 785.99 \subseteq \perp $ \\
   \hline
\end{tabular}
\end{table}

Table~\ref{table:iteration} lists all the steps performed by the
 \textsc{CounterFactualXplain} algorithm to find a counterfactual explanation and a maximal relaxation to the given problem with the set of constraints $\mathcal{B} = \{c_5\}$ and $\mathcal{F} = \{c_1, c_2, c_3, c_4\}$.
Note that the given set of constraints $\mathcal{C} = \mathcal{B} \cup \mathcal{F}$ is inconsistent.
The algorithm initializes the set of constraints $C' = \{ \top_1, \top_2, \top_3, \top_4 \}$.
Let us introduce subsets of constraints denoted by $S_i$ to represent the elements in 
 $C'$ at each iteration.
The initial set is $S_0 = C'$, and
 the subsequent subsets are identified by the iteration number in the table and are accumulated as $S_i = S_{i-1} \cup r_j$, where $r_j$ denotes the next tightening performed on the constraints.

\begin{table}[t]
\caption{The list of all iterations performed by the \textsc{CounterFactualXplain} to find a counterfactual explanation given the background constraint set $\mathcal{B} = \{price \leq 2000\}$.}
\label{table:iteration} 
\centering
\footnotesize
\begin{tabular}{|clcc|}
\hline
$i$ & Subset ($S_i$) & $S_i$ consistent? & $\mathcal{E}$ \\ \hline
$1$ & $S_1 = S_0 \cup \{ c_1$ = `Lenovo'$\}$ & true & \{\} \\\hline
$2$ & $S_2 = S_1 \cup \{ c_2 \geq 11.1 \}$ & true & \{\} \\\hline
$3$ & $S_3 = S_2 \cup \{ c_2 \geq 14.0 \}$ & true & \{\} \\\hline
$4$ & $S_4 = S_3 \cup \{ c_2 \geq 15.0 \}$ & true & \{\} \\\hline
$5$ & $S_5 = S_4 \cup \{ c_3 \geq 512 \}$ & true & \{\} \\\hline
$6$ & $S_6 = S_5 \cup \{ c_4 \geq 2.2 \}$ & true & \{\} \\\hline
$7$ & $S_7 = S_6 \cup \{ c_4 \geq 4.5 \}$ & false & \{$c_4 \geq 2.2$\} \\\hline
\end{tabular}
\end{table}
%\bccom{I did not really like this example. I would love to find a scenario with a few inconsistencies.}

In Table~\ref{table:iteration}, the first iteration tightens $c_1$ to `Lenovo', which corresponds to the initial user constraint $c_1$, and the set of constraints corresponding to this iteration $S_1$ is consistent.
Therefore, in the next iterations (from $2$ to $4$ inclusive), the constraint
 tightening is performed for the next constraint $c_2$.
As it is possible to tighten the $c_2$ until the original user constraint,
 the fifth iteration, tightens the next constraint, i.e. $c_3$.
Similarly, iterations $6$ and $7$ performs tightening on $c_4$, where the seventh iteration
 with $c_4 \geq 4.5$ makes the set of constraints inconsistent. 
Therefore, the tightest version of this constraint that is consistent is added to
 the explanation. 
Finally, the algorithm returns the maximal relaxation $C' = S_6$, and the
 counterfactual explanation $\mathcal{E} = \{ c_4 \geq 2.2 \}$. 
The user-interface can inform the user with an explanation that is similar
 to: ``If you change your constraint on battery life from 10 hr to 2.2 hr, you can find at least one solution that satisfies your remaining constraints''.
The relaxed CSP $\phi' = \langle \mathcal{X}, \mathcal{D}, C' \rangle$
 contains a single solution, which is \{Lenovo, 15.4 inches, 1024.0 MB, 2.2 hr, \$1499.99\}.

It is important to note here, one can argue that the item
 \{Lenovo, 14.0 inches, 512 MB, 4.5 hr, \$1899\} is closer to the initial solution
 than the solution found by our approach by applying another metric.
Our aim in this paper is to find a set of changes that can be applied to the system
 to change the outcome (feasibility state) of the system. 
At this stage, we discuss only preliminary research findings, and the
 relation between system-based minimal changes vs. solution-based minimal changes needs
 to be studied further.

\section{Discussion and Future Work}

We propose a novel explanation type for constraint based systems by using the
 counterfactual explanation framework and identifying a maximal relaxation of the constraint set. 
Our proposed notion of counterfactual explanations aims to find a minimal set of changes 
 for the set of user constraints using multi-point relaxation spaces that allows the user
 to find a solution. 
However, an important point to note here is that a minimal perturbation in the 
 constraint set may not necessarily lead to minimal changes on the solution that was first presented to the user.
As future work, we intend to further investigate the relation between minimal
 changes on the set of constraints and its effect on the solution. 
Our plan includes conducting a user study to also understand the social impact of it.

%We intend to take the proposed algorithm and implement it using real world scenarios and various data sets. We also intend to build an algorithm from a solution based perspective instead of the system based one. Using these algorithms and applying them to configuration systems we can analyse the performance and issues that come up. We also intend on applying an interactive tool using the solution based approach and do an in depth analysis of the user case studies. 

\section{Acknowledgments}
This publication has emanated from research conducted with the financial support of Science Foundation Ireland under Grant 16/RC/3918, 12/RC/2289-P2, and 18/CRT/6223, which are co-funded under the European Regional Development Fund.
This research was also partially supported by TAILOR, HumanE AI Network, BRAINE, and StairwAI projects funded by EU Horizon 2020 under Grant Agreements 952215, 952026, 876967, and 101017142.

\bibliography{aaai22}
\label{sec:reference_examples}

\end{document}